\pdfoutput=1
\documentclass[10pt]{article}
\usepackage{ml_preprint}

\title{The Missing Root: Geometric Imagination\\Without Images in Large Language Models}

\author{%
  \textbf{Lo\"ic Magnan} \qquad \textbf{Tomasz Kozlowski} \\[4pt]
  The Grainger College of Engineering \\
  University of Illinois Urbana-Champaign \\
  \texttt{\{lmagnan2, txk\}@illinois.edu}
}

\begin{document}
\maketitle

\begin{abstract}
\noindent
This paper proposes a new paradigm and method for giving large language models spatial understanding. Standard language models operate exclusively over discrete words (secondary representations of things rather than continuous space itself), forcing them to guess geometry verbally. Whereas most multimodal approaches equate spatial grounding with camera images or video frames, spatial structure itself is not a picture: it consists of coordinate frames, solid boundaries, and how objects move and relate to each other in continuous space. Motivated by this distinction, we place continuous object tokens combining pose (position, orientation, and scale) and signed-distance shape embeddings alongside word tokens inside a single uninterrupted transformer sequence, allowing language and geometry to attend to each other as object scenes are built and updated step by step. Control tokens switch output automatically between natural-language words and continuous object state updates to a persistent object workspace, while an attention mask blocks earlier prose when generating the answer, so responses must be read directly from settled geometry. To prove this method works, we attach a narrow parallel spatial stream ($+4.4\%$ weights) alongside unfrozen \texttt{Qwen3-1.7B} layers. In our two-dimensional tests, the model spawns and transforms multi-object scenes from English descriptions, transfers shapes to held-out nouns, and relies causally on the workspace to answer spatial questions.
\end{abstract}

% ============================================================================
\section{Introduction}
% ============================================================================

Scaling text transformers~\cite{vaswani2017attention} has validated a surprising bet about language: training a large language model (LLM) on trillions of tokens of prose, mathematics, and code recovers far more of the causal structure of the world than early critics expected. Yet in \emph{On the Fourfold Root of the Principle of Sufficient Reason}~\cite{schopenhauer1813fourfold,schopenhauer1818world}, Arthur Schopenhauer drew a sharp distinction between two very different foundations of thought. Conceptual reason, the \emph{root of knowing}, operates over words, which remain \emph{``representations of representations''}: discrete, compressed pointers to things rather than the things themselves. Underneath language sits a separate foundation, the \emph{principle of sufficient reason of being}: the direct intuition of how objects occupy, relate to, and constrain one another in continuous space and time. Standard language models possess only the first root. Because a text transformer processes everything as a one-dimensional chain of discrete symbols, it holds no internal mechanism for continuous form, orientation, or clearance. Even when strong verbal fluency or external software harnesses compensate for this absence in practice, the model itself never directly represents the space its words describe.

In living minds, this spatial root is not learned from text alone; it is built from direct interaction with the physical world through channels such as sight and touch. Because vision is so dominant in everyday human life, most efforts to ground LLMs assume that spatial understanding requires visual pictures, attaching camera patch encoders in vision-language models~\cite{radford2021clip} or rendering pixel frames in video generators. Human neuroscience shows that the internal spatial sense itself does not require an image. People with congenital \emph{aphantasia} cannot form voluntary visual pictures in their heads at all, yet match normal accuracy on mental rotation and spatial layout tasks~\cite{zeman2015aphantasia,bainbridge2021quantifying,pounder2022minimal}. Likewise, individuals born completely blind develop accurate spatial intuition through hands, motion, and surface contact~\cite{ricciardi2014supramodal}. In both cases, the brain's dorsal \emph{parietal cortex}~\cite{goodale1992separate} maintains a non-visual representation of coordinate frames, solid boundaries, and spatial transformations without painting an internal image. Giving an LLM the missing root of space, therefore, should not require giving it eyes. We postulate it is sufficient (and more appropriate) to give it an internal, non-visual geometric state (closer to mathematical touch and object manipulation than to photography) in which solid shapes and poses can be held and moved alongside words.

Viewing spatial grounding as non-visual state manipulation sets our approach apart from existing paradigms. Vision-language models and pixel world models process two-dimensional camera patches: contrastive encoders discard exact metric pose in favor of caption-level features~\cite{radford2021clip}, interiors remain hidden behind outer surfaces, and, when designing or reasoning from prose, no input photograph exists to encode in the first place. Vision-language-action robotics models such as $\pi_0$~\cite{black2024pi0} attach continuous output heads, but operate one-way from live camera images out to motor commands rather than closing an internal loop from words to persistent objects and back to words. Conversely, latent-thought language models such as Coconut~\cite{hao2024coconut} pass unconstrained hidden vectors between steps without enforcing metric coordinates or solid boundaries, while external code tool-calling leaves the language model itself blind while describing geometry.

At a high level, our approach of coupling a pretrained language model with continuous spatial weights and an object head enables more than either component could achieve alone. Even without direct spatial intuition, a language model trained on trillions of words already stores vast implicit knowledge about the physical world inside its weights: it knows what everyday objects do, that a banana curves, a ruler is straight, and how containers, supports, and mechanical tools relate in human activity. What it lacks is a continuous medium where positions, orientations, and solid boundaries persist rather than being approximated in words. Placing continuous object tokens directly into the same sequence as words, paired with an object head that writes pose and shape updates to a persistent workspace, lets language weights and spatial weights exchange information across layers. Language weights supply broad world knowledge so the model can infer plausible shapes and relative layouts for unfamiliar nouns without needing millions of labeled meshes, while the spatial stream and workspace enforce metric coordinates and boundaries that the language model can read back to answer questions. Each side covers the other's weakness: language brings knowledge of what things are, and the workspace provides the continuous space to represent them.

In this work, we present a series of experiments designed to test this idea. We introduce a spatial workspace paradigm in which continuous object tokens sit alongside word tokens inside a single transformer sequence (Section~\ref{sec:paradigm}). We achieve this by adding two-dimensional object representation and manipulation to a pretrained 1.7-billion-parameter language model (\texttt{Qwen3-1.7B}~\cite{yang2025qwen3}, Section~\ref{sec:grafting}). With less than five percent added weights and no external geometry tools at inference, the model spawns multi-object scenes from plain English descriptions (even when no coordinates are given), moves and rotates shapes across multiple steps, and answers questions about containment, proximity, and relative position directly from the final scene (Section~\ref{sec:results}). Most notably, training our parallel spatial stream on $2{,}840$ hand-labeled noun-to-form pairs over roughly fifty basic shape silhouettes draws out physical shape structure already latent in pretrained words, allowing held-out nouns omitted from spatial training (such as an \emph{umbrella}, a \emph{croissant}, or a \emph{crayon}) to materialize with accurate shapes on the object workspace.

% ============================================================================
\section{The Spatial Workspace Paradigm}
\label{sec:paradigm}
% ============================================================================

\begin{figure}[!ht]
  \centering
  \resizebox{\textwidth}{!}{\begin{tikzpicture}[
    font=\sffamily,
    >= {Stealth[length=4.4pt, width=3.5pt]},
    line cap=round, line join=round,
    thinarr/.style={->, draw=muted, line width=0.6pt},
    bluearr/.style={->, draw=llmstroke, line width=0.65pt},
    greenarr/.style={->, draw=geostroke, line width=0.7pt}
]

\fill[panelbg, rounded corners=6pt] (0, 0) rectangle (17.28, 6.05);

% ============================================================================
% 1A. WORD LANE (TOP, y = 4.58) -- Aligned column-by-column with Object Lane
% ============================================================================
% Col 1: Raw word [0.30 .. 1.06]
\draw[draw=llmstroke, line width=0.6pt, fill=white, rounded corners=2.5pt]
    (0.30, 4.23) rectangle (1.06, 4.93);
\node[font=\sffamily\scriptsize, text=ink] at (0.68, 4.58) {Word};

\draw[bluearr] (1.10, 4.58) -- (1.36, 4.58);

% Col 2: Tokenizer [1.40 .. 2.56] (aligned with Encoder)
\draw[draw=llmstroke, line width=0.6pt, fill=white, rounded corners=2.5pt]
    (1.40, 4.23) rectangle (2.56, 4.93);
\node[font=\sffamily\scriptsize, text=ink] at (1.98, 4.58) {Tokenizer};

\draw[bluearr] (2.60, 4.58) -- (2.86, 4.58);

% Col 3: Discrete token ID [2.90 .. 3.48] (aligned with [pose; z_32])
\draw[draw=llmstroke, line width=0.6pt, fill=tokword, rounded corners=1.8pt]
    (2.90, 4.23) rectangle (3.48, 4.93);
\node[font=\sffamily\tiny, align=center, text=ink] at (3.19, 4.58) {token\\[-1pt]ID};

\draw[bluearr] (3.52, 4.58) -- (3.92, 4.58);

% Col 4: Embedding Table / Matrix [3.96 .. 5.12] (aligned with W_in)
\draw[draw=llmstroke, line width=0.65pt, fill=llmfill, rounded corners=2.5pt]
    (3.96, 3.78) rectangle (5.12, 5.38);
\foreach \ry in {3.98, 4.14, 5.02, 5.18} {
    \draw[draw=llmstroke!32, line width=0.4pt] (4.10, \ry) -- (4.98, \ry);
}
\node[font=\scriptsize, align=center, text=ink] at (4.54, 4.58)
    {$W_{\text{emb}}$\\[1pt]{\sffamily\tiny\color{llmstroke}embed table}};

\draw[bluearr] (5.16, 4.58) -- (5.52, 4.58);

% Col 5: Sequence Word Vector [5.56 .. 5.98] (aligned with Object Vector)
\draw[draw=llmstroke, line width=0.6pt, fill=tokword, rounded corners=1.8pt]
    (5.56, 4.00) rectangle (5.98, 5.16);
\foreach \y in {4.20, 4.45, 4.71, 4.96} {
    \fill[llmstroke] (5.77, \y) circle (0.7pt);
}
\node[font=\scriptsize, text=llmstroke] at (5.77, 5.40) {$\vec{e}_w$};

\draw[bluearr] (6.02, 4.58) -- (6.60, 4.58);

% ============================================================================
% 1B. OBJECT LANE (BOTTOM, y = 2.48) -- Aligned column-by-column with Word Lane
% ============================================================================
% Col 1: Shape outlines centered around Encoder [y = 2.02]
\draw[draw=geostroke, line width=0.7pt, fill=tokobj]
    (0.44, 2.86) -- (0.44, 2.44) -- (0.92, 2.44) -- (0.92, 2.86)
    -- (0.80, 2.86) -- (0.80, 2.55) -- (0.56, 2.55) -- (0.56, 2.86) -- cycle;
\draw[draw=geostroke, line width=0.7pt, fill=tokobj]
    (0.45, 1.84) -- (0.91, 1.84) -- (0.68, 2.24) -- cycle;
\draw[draw=geostroke, line width=0.7pt, fill=tokobj]
    (0.68, 1.42) circle (0.20cm);
\node[font=\sffamily\tiny, text=muted] at (0.68, 1.02) {outlines};

\draw[greenarr] (1.04, 2.02) -- (1.36, 2.02);

% Col 2: Encoder [1.40 .. 2.56] (aligned with Tokenizer)
\draw[draw=geostroke, line width=0.6pt, fill=white, rounded corners=2.5pt]
    (1.40, 1.67) rectangle (2.56, 2.37);
\node[font=\sffamily\scriptsize, text=ink] at (1.98, 2.02) {Encoder};

\draw[greenarr] (2.60, 2.02) -- (2.86, 2.02);

% Col 3: Pose + Shape Embedding [2.90 .. 3.48] (aligned with token ID)
\draw[draw=geostroke, line width=0.6pt, fill=white, rounded corners=1.8pt]
    (2.90, 2.58) rectangle (3.48, 3.42);
\node[font=\tiny, align=center, text=ink] at (3.19, 3.00) {$x$\\[-1pt]$y$\\[-1pt]$\theta, s$};
\node[font=\sffamily\tiny, text=muted, anchor=east] at (2.86, 3.00) {pose};

\draw[draw=geostroke, line width=0.6pt, fill=tokobj, rounded corners=1.8pt]
    (2.90, 1.54) rectangle (3.48, 2.48);
\node[font=\scriptsize, text=geostroke] at (3.19, 2.01) {$\vec{z}_{32}$};

\node[font=\scriptsize, text=muted] at (3.72, 2.48) {$\times$};

% Col 4: Input Matrix W_in [3.96 .. 5.12] (aligned with W_emb)
\draw[draw=geostroke, line width=0.65pt, fill=geofill, rounded corners=2.5pt]
    (3.96, 1.68) rectangle (5.12, 3.28);
\node[font=\scriptsize, align=center, text=ink] at (4.54, 2.48)
    {$W_{\text{in}}$\\[1pt]{\sffamily\tiny\color{geostroke}input matrix}};

\draw[greenarr] (5.16, 2.48) -- (5.52, 2.48);

% Col 5: Sequence Object Vector [5.56 .. 5.98] (aligned with Word Vector)
\draw[draw=geostroke, line width=0.6pt, fill=tokobj, rounded corners=1.8pt]
    (5.56, 1.90) rectangle (5.98, 3.06);
\foreach \y in {2.10, 2.35, 2.61, 2.86} {
    \fill[geostroke] (5.77, \y) circle (0.7pt);
}
\node[font=\scriptsize, text=geostroke] at (5.77, 3.30) {$\vec{e}_{\text{obj}}$};

\draw[greenarr] (6.02, 2.48) -- (6.60, 2.48);

% ============================================================================
% 2. LANGUAGE MODEL + PERSISTENT OBJECT WORKSPACE
% ============================================================================
\draw[draw=llmstroke, line width=0.75pt, fill=llmfill, rounded corners=4pt]
    (6.64, 1.98) rectangle (9.80, 5.08);
\node[font=\sffamily\small, align=center, text=ink] at (8.22, 3.74) {Language Model};
\node[font=\sffamily\tiny, text=llmstroke] at (8.22, 3.22) {words \& objects share sequence};

\draw[draw=geostroke, line width=0.65pt, fill=white, rounded corners=3.5pt]
    (6.74, 0.42) rectangle (9.68, 1.72);
\node[font=\sffamily\scriptsize, text=ink] at (8.21, 1.47) {Object Workspace};

\draw[draw=geostroke, line width=0.55pt, fill=tokobj, rounded corners=1.8pt]
    (7.15, 0.58) rectangle (7.98, 1.20);
\node[font=\sffamily\tiny, text=ink] at (7.56, 0.89) {triangle};

\draw[draw=geostroke, line width=0.55pt, fill=tokobj, rounded corners=1.8pt]
    (8.42, 0.58) rectangle (9.25, 1.20);
\node[font=\sffamily\tiny, text=ink] at (8.83, 0.89) {cup};

\draw[greenarr, densely dashed]
    (6.74, 0.96) .. controls (5.40, 0.72) and (3.72, 0.76) .. (3.19, 1.50);
\node[font=\sffamily\tiny, text=geostroke] at (4.98, 0.54) {re-read on \texttt{<STEP>} / \texttt{<LOOK>}};

% ============================================================================
% 3. DUAL HEADS (Aligned horizontally with Word & Object lanes)
% ============================================================================
\draw[bluearr]  (9.80, 4.58) -- (10.30, 4.58);
\draw[greenarr] (9.80, 2.48) -- (10.30, 2.48);

\draw[draw=llmstroke, line width=0.65pt, fill=tokword, rounded corners=3pt]
    (10.34, 4.08) rectangle (12.18, 5.08);
\node[font=\sffamily\scriptsize, align=center, text=ink] at (11.26, 4.58)
    {Word Head\\[1pt]{\tiny\color{llmstroke}words \& control tokens}};

\draw[draw=geostroke, line width=0.65pt, fill=geofill, rounded corners=3pt]
    (10.34, 1.98) rectangle (12.18, 2.98);
\node[font=\sffamily\scriptsize, align=center, text=ink] at (11.26, 2.48)
    {Object Head\\[1pt]{\tiny\color{geostroke}$\Delta\vec{p},\, \Delta\vec{z}$}};

\draw[thinarr, densely dashed] (11.26, 4.08) -- (11.26, 2.98)
    node[midway, right, font=\ttfamily\tiny, text=muted] {<NEW>/<STEP>};

\draw[greenarr]
    (11.26, 1.98) .. controls (11.26, 1.12) and (10.45, 0.96) .. (9.72, 0.96);
\node[font=\scriptsize, text=geostroke] at (11.38, 0.82) {$\vec{p} \leftarrow \vec{p} + \Delta\vec{p}$};

% ============================================================================
% 4. EXAMPLE SEQUENCE CARD
% ============================================================================
\draw[draw=muted!45, line width=0.55pt, fill=white, rounded corners=3.5pt]
    (12.68, 0.42) rectangle (16.92, 5.38);
\node[anchor=north west, font=\sffamily\bfseries\scriptsize, text=ink] at (12.92, 5.14) {Example sequence};

\node[anchor=north west, font=\sffamily\tiny, text width=3.85cm, text=muted, inner sep=0pt] at (12.92, 4.66)
    {\textbf{Prompt:} ``Triangle at $(-0.6, 0)$, cup at $(0.2, 0)$. Move triangle $0.8$ right. Is it in the cup?''};

\draw[draw=muted!30, line width=0.4pt] (12.92, 3.60) -- (16.68, 3.60);

\node[anchor=north west, font=\ttfamily\tiny, text width=3.85cm, text=ink, inner sep=0pt] at (12.92, 3.40)
    {1. <NEW> triangle <NEW> cup <STEP>};
\node[anchor=north west, font=\sffamily\tiny, text=geostroke, inner sep=0pt] at (13.12, 3.04)
    {$\rightarrow$ spawns $[\vec{p}_1, \vec{z}_1],\, [\vec{p}_2, \vec{z}_2]$};

\node[anchor=north west, font=\ttfamily\tiny, text width=3.85cm, text=ink, inner sep=0pt] at (12.92, 2.50)
    {2. "Move triangle 0.8 right." <STEP>};
\node[anchor=north west, font=\sffamily\tiny, text=geostroke, inner sep=0pt] at (13.12, 2.14)
    {$\rightarrow$ updates $\Delta x = +0.8$ in workspace};

\node[anchor=north west, font=\ttfamily\tiny, text width=3.85cm, text=ink, inner sep=0pt] at (12.92, 1.60)
    {3. <LOOK> \textrm{\color{muted}(masks prior text)}};
\node[anchor=north west, font=\sffamily\tiny, text=llmstroke, inner sep=0pt] at (13.12, 1.24)
    {$\rightarrow$ reads settled objects $\rightarrow$ \texttt{\textbf{"yes"}}};

\end{tikzpicture}}
  \caption{\textbf{Overview of the spatial workspace paradigm.} Words and continuous objects share a single sequence through the language model (treated as a black box). Each object combines a $5$-element pose $(x, y, \cos\theta, \sin\theta, s)$ with a $32$-element signed-distance shape embedding $\vec{z}_{32}$ compressed from its $128$-point outline~\cite{park2019deepsdf}, projected by $W_{\text{in}}$ into object embeddings $\vec{e}_{\text{obj}}$ alongside word embeddings $\vec{e}_w$ projected by $W_{\text{emb}}$. Control tokens switch between natural language (Word Head) and object representation (Object Head). Emitting the \texttt{<LOOK>} token masks the entire prior sequence so the model can only read the current geometry from the object workspace.}
  \label{fig:overview}
\end{figure}
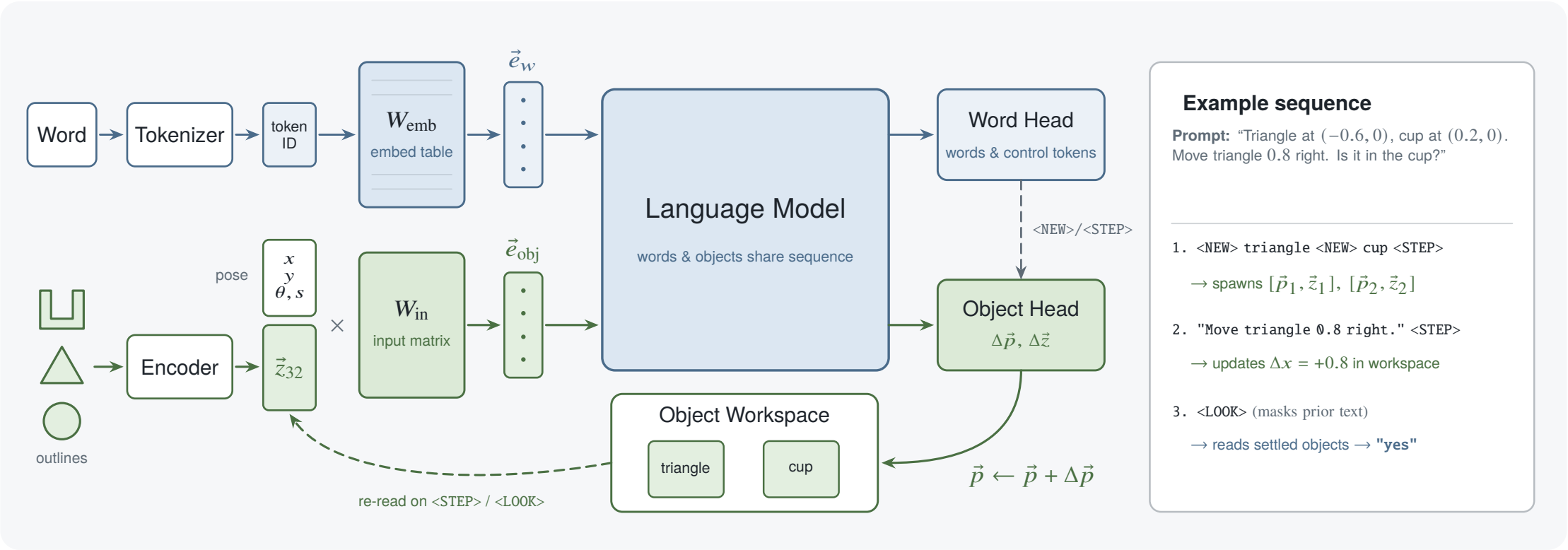

First, we describe the workspace paradigm while treating the language model as a black box (Figure~\ref{fig:overview}). We define the input vectors, output heads, control tokens, and external workspace, and then show how they execute together inside a single inference sequence.

\subsection{Object Tokens and Shape Encoding}
A transformer layer does not inherently require its inputs to be words; it operates over a sequence of vectors in a shared embedding space. Although standard language models look up every token in a discrete vocabulary table, a token can carry any structured state projected into that space (such as three-dimensional objects, velocities, forces, material properties, or other physical quantities), allowing words and non-verbal entities to attend to each other. Here we instantiate this general idea with two-dimensional rigid objects.

Each object in our workspace separates its \emph{pose} from its \emph{shape}. Its pose is stored as a five-element vector $\vec{p} = (x, y, \cos\theta, \sin\theta, s) \in \mathbb{R}^5$, corresponding to center coordinates $(x, y)$, rotation $(\cos\theta, \sin\theta)$ (written as cosine and sine rather than a raw angle, so rotation wraps smoothly at $0^\circ \equiv 360^\circ$), and size $s$. To help linear projections resolve fine differences in position, $(x, y, s)$ are also expanded across six fixed geometric sine-and-cosine spatial frequencies $\gamma(x, y, s) \in \mathbb{R}^{36}$.

Independently of pose, an object's intrinsic geometry is represented by a compact shape embedding $\vec{z}_{32} \in \mathbb{R}^{32}$ produced by a small, separately trained and frozen boundary autoencoder (\textsf{Encoder} in Figure~\ref{fig:overview}). Instead of pixel images, the encoder takes a closed loop of $128$ evenly spaced $(x, y)$ points around the perimeter of a centered, unit-radius shape and passes them through four one-dimensional convolution layers that wrap around the closed perimeter, so the resulting embedding does not depend on which perimeter point was listed first. During autoencoder training, a paired signed-distance decoder~\cite{park2019deepsdf} checks that $\vec{z}_{32}$ captures solid geometry by predicting the signed distance $D(\vec{z}_{32}, \vec{q})$ from any query point $\vec{q} \in \mathbb{R}^2$ to the boundary (negative inside the object, zero on the surface, positive outside the object). Together, the pose, Fourier-like position expansion, and shape embedding form a $73$-element vector $\vec{f}$, which an input projection $W_{\text{in}}$ (with learned bias vector $\vec{b}_{\text{in}}$) maps into an object embedding $\vec{e}_{\text{obj}}$ that enters the sequence alongside word embeddings $\vec{e}_w$:
\begin{equation}
  \vec{f} = \big[\,\vec{p},\; \gamma(x, y, s),\; \vec{z}_{32}\,\big] \in \mathbb{R}^{73}, \qquad \vec{e}_{\text{obj}} = W_{\text{in}}\vec{f} + \vec{b}_{\text{in}}.
  \label{eq:obj_in}
\end{equation}

\subsection{The Object Workspace, Control Tokens, and Output Heads}
At the output of the language model sit two output heads connected to an external memory array, the \emph{object workspace} (center-right and bottom of Figure~\ref{fig:overview}).

\emph{Object workspace.} The object workspace is an external memory array of size $K \times 37$ that stores the current state $[\vec{p}_k, \vec{z}_k]$ ($5$-element pose $\vec{p}_k$ and $32$-element shape embedding $\vec{z}_k$) for each of the $K$ active objects in the scene. It performs no geometric computation of its own; it simply preserves these floating-point states across steps so multi-step edits do not lose precision through repeated conversion to text, and feeds them back into $W_{\text{in}}$ (recomputing the $36$-element Fourier position expansion on entry) whenever an update or final check is triggered.

\emph{Word head and control tokens.} On word tokens, the model's standard vocabulary head is extended with five special control tokens:
\begin{itemize}[leftmargin=1.5em, itemsep=1pt, topsep=2pt]
  \item \texttt{<SCENE>}: introduces objects provided directly at the start of a problem;
  \item \texttt{<NEW>}: declares a new object described by the words that follow, adding an empty (all-zero) entry to the object workspace;
  \item \texttt{<STEP>} and \texttt{</STEP>}: open and close a state-update pass over all active objects in the workspace; and
  \item \texttt{<LOOK>}: appends the current object workspace ahead of the question and masks all earlier text so the answer must be read from the current objects.
\end{itemize}
Because ordinary words and control tokens share a single softmax distribution, the model switches between natural language and workspace operations within a single pass, with no external conditionals or hardcoded branching.

\emph{Object head.} Whenever \texttt{<STEP>} opens an update pass, the active entries in the object workspace are appended to the sequence as object tokens. On each object token, the object head outputs a $37$-element update vector $(\Delta\vec{p}, \Delta\vec{z}_{32}) \in \mathbb{R}^{5 + 32}$ that updates that object's stored state:
\begin{equation}
  \vec{p} \leftarrow \vec{p} + \Delta\vec{p}, \qquad \vec{z}_{32} \leftarrow \vec{z}_{32} + \Delta\vec{z}_{32},
  \label{eq:delta}
\end{equation}
where $\vec{p}$ is the object's pose and $\vec{z}_{32}$ is its shape embedding. Using additive updates gives one shared rule for both object creation and rigid motion: a newly declared object starts at $(\vec{p}, \vec{z}_{32}) = (\vec{0}, \vec{0})$, so its first \texttt{<STEP>} update supplies its initial pose and shape; when an existing object is translated or rotated, training targets $\Delta\vec{z}_{32} = \vec{0}$ alongside the target $\Delta\vec{p}$ so rigid motion leaves its shape unchanged.

\subsection{Sequence Execution and Head Switching}
Once prompted in natural language, the entire problem (reading the prompt, creating the described objects, manipulating them according to the instructions, and answering spatial questions) runs as one uninterrupted sequence rather than re-prompting the model across a multi-turn conversation. Word embeddings and object embeddings occupy adjacent tokens within a single sequence, and the emitted control tokens automatically switch which output head is active at each token. This is similar to how \texttt{BOS}/\texttt{EOS} and other special tokens work, triggering desired changes in inference behavior.

Following the example on the right of Figure~\ref{fig:overview}, when the prompt introduces a triangle and a cup, the word head emits a \texttt{<NEW>} token followed by each object's name, registering blank placeholder entries in the object workspace. When the word head emits a \texttt{<STEP>} token, the workspace entries (representing physical objects) are appended to the end of the active sequence as object tokens, and output on those tokens switches automatically to the object head to populate $[\vec{p}_1, \vec{z}_1]$ and $[\vec{p}_2, \vec{z}_2]$. Objects in the same step share a single position index and attend to each other bidirectionally so they are treated as an unordered set rather than a left-to-right list. The workspace entries are updated with these new states and appended to the sequence after the \texttt{</STEP>} token (so subsequent tokens can read the updated scene), returning control to the word head. Subsequent object-manipulation instructions (such as move or rotate) trigger additional \texttt{<STEP>} tokens, which update the object tokens as described above. Finally, the word head emits a \texttt{<LOOK>} token, which appends the current workspace objects and the user's question so the word head generates the natural-language answer to the prompt (e.g., position, distance, or composition of the objects).

Opening and updating the object workspace is not intended to be the LLM's default state. As long as no control tokens are emitted, inference continues as an ordinary language pass through the model, leaving non-spatial conversation and reasoning unchanged.

\subsection{Preventing Verbal Shortcuts with the \texttt{<LOOK>} Mask}
Combining word and object embeddings inside one sequence raises a dual-representation problem. Suppose a prompt places a triangle at $(-0.6, 0)$ and a cup at $(0.2, 0)$, instructs the model to move the triangle $0.8$ units right, and asks whether the triangle ends up inside the cup. Under standard left-to-right causal attention, the new (inferred) tokens see the original prompt with its natural-language object position description, allowing a capable language model to infer $-0.6 + 0.8 = 0.2$ verbally while ignoring the object tokens altogether.

\begin{figure}[!ht]
  \centering
  \resizebox{\textwidth}{!}{\begin{tikzpicture}[
    font=\sffamily,
    >= {Stealth[length=4.4pt, width=3.5pt]},
    line cap=round, line join=round,
    bluearr/.style={->, draw=llmstroke, line width=0.65pt},
    greenarr/.style={->, draw=geostroke, line width=0.65pt}
]

% PANEL (a): INSIDE THE TRANSFORMER
\fill[panelbg, rounded corners=6pt] (0, 0) rectangle (9.85, 5.75);
\node[anchor=north west, font=\sffamily\bfseries\footnotesize, text=ink] at (0.28, 5.52)
    {(a) Inside the transformer: parallel spatial stream ($\times 28$ layers)};

\draw[draw=llmstroke, line width=0.6pt, fill=tokword, rounded corners=2pt]
    (0.48, 0.62) rectangle (3.92, 1.02);
\node[font=\scriptsize, text=ink] at (2.20, 0.82) {Language $\vec{h}^{(\ell)} \in \mathbb{R}^{2048}$};

\draw[draw=geostroke, line width=0.6pt, fill=tokobj, rounded corners=2pt]
    (5.62, 0.62) rectangle (9.35, 1.02);
\node[font=\scriptsize, text=ink] at (7.48, 0.82) {Geometry $\vec{g}^{(\ell)} \in \mathbb{R}^{256}$};

\draw[draw=llmstroke, line width=0.7pt, fill=llmfill, rounded corners=3.5pt]
    (0.48, 1.78) rectangle (3.92, 3.92);
\node[font=\sffamily\footnotesize, align=center, text=ink] at (2.20, 3.05)
    {Pretrained LLM Layer\\[2pt]{\scriptsize\color{llmstroke}Qwen3-1.7B ($16$ heads)}};
\node[font=\sffamily\tiny, text=muted] at (2.20, 2.22) {frozen $\rightarrow$ $3\!\times\!10^{-6}$ LR};

\draw[draw=geostroke, line width=0.7pt, fill=geofill, rounded corners=3.5pt]
    (5.62, 1.78) rectangle (9.35, 3.92);
\node[font=\sffamily\footnotesize, align=center, text=ink] at (7.48, 3.22)
    {Parallel Spatial Block};
\node[font=\sffamily\tiny, align=center, text=geostroke] at (7.48, 2.42)
    {$2$ QK-Norm heads $\times 128$ + MLP\\[2pt]\color{muted}$[256\text{ lang} \;\|\; 256\text{ geo}]$ (equal width)};

\draw[bluearr]  (2.20, 1.02) -- (2.20, 1.78);
\draw[greenarr] (7.48, 1.02) -- (7.48, 1.78);
\draw[bluearr]  (2.20, 3.92) -- (2.20, 4.54);
\draw[greenarr] (7.48, 3.92) -- (7.48, 4.54);

\draw[bluearr] (3.92, 2.42) -- (5.62, 2.42);
\node[font=\sffamily\tiny, text=ink] at (4.77, 2.64) {down-proj};
\node[font=\tiny, text=muted] at (4.77, 2.18) {$2048 \!\rightarrow\! 256$};

\draw[greenarr, densely dashed] (5.62, 3.38) -- (3.92, 3.38);
\node[font=\tiny, text=geostroke] at (4.77, 3.60) {$B\!=\!0$ cross-proj};
\node[font=\sffamily\tiny, text=muted] at (4.77, 3.14) {layers $6\text{--}27$};

\draw[draw=llmstroke, line width=0.6pt, fill=tokword, rounded corners=2pt]
    (0.48, 4.54) rectangle (3.92, 4.94);
\node[font=\scriptsize, text=ink] at (2.20, 4.74) {$\vec{h}^{(\ell+1)} \in \mathbb{R}^{2048}$};

\draw[draw=geostroke, line width=0.6pt, fill=tokobj, rounded corners=2pt]
    (5.62, 4.54) rectangle (9.35, 4.94);
\node[font=\scriptsize, text=ink] at (7.48, 4.74) {$\vec{g}^{(\ell+1)} \in \mathbb{R}^{256}$};

% PANEL (b): ATTENTION VISIBILITY MASK
\fill[panelbg, rounded corners=6pt] (10.18, 0) rectangle (17.28, 5.75);
\node[anchor=north west, font=\sffamily\bfseries\footnotesize, text=ink] at (10.46, 5.52)
    {(b) Attention visibility mask};

\def\mx{12.45}
\def\my{0.72}
\def\cs{0.88}

\foreach \r in {0,1,2,3} {
    \foreach \c in {0,1,2,3} {
        \pgfmathsetmacro{\x}{\mx + \c*\cs}
        \pgfmathsetmacro{\y}{\my + (3-\r)*\cs}
        \fill[blocked] (\x, \y) rectangle (\x+\cs, \y+\cs);
    }
}

\fill[llmfill] (\mx, \my+3*\cs) -- (\mx, \my+4*\cs) -- (\mx+\cs, \my+3*\cs) -- cycle;
\fill[llmfill] (\mx, \my+2*\cs) rectangle (\mx+\cs, \my+3*\cs);
\fill[geofill] (\mx+\cs, \my+2*\cs) rectangle (\mx+2*\cs, \my+3*\cs);
\fill[geofill] (\mx+2*\cs, \my+\cs) rectangle (\mx+3*\cs, \my+2*\cs);
\fill[geofill] (\mx+2*\cs, \my) rectangle (\mx+3*\cs, \my+\cs);
\fill[llmfill] (\mx+3*\cs, \my) -- (\mx+3*\cs, \my+\cs) -- (\mx+4*\cs, \my) -- cycle;

\foreach \i in {0,1,2,3,4} {
    \draw[draw=muted!60, line width=0.45pt] (\mx+\i*\cs, \my) -- (\mx+\i*\cs, \my+4*\cs);
    \draw[draw=muted!60, line width=0.45pt] (\mx, \my+\i*\cs) -- (\mx+4*\cs, \my+\i*\cs);
}

\draw[draw=barrier, line width=0.8pt, densely dashed]
    (\mx, \my) rectangle (\mx+2*\cs, \my+2*\cs);
\node[font=\sffamily\tiny, align=center, text=barrier] at (\mx+\cs, \my+\cs)
    {blocked\\past \texttt{<LOOK>}};

\node[font=\sffamily\tiny, text=geostroke] at (\mx+1.5*\cs, \my+2.5*\cs) {bi-dir};
\node[font=\sffamily\tiny, text=geostroke] at (\mx+2.5*\cs, \my+1.5*\cs) {bi-dir};

\node[anchor=east, font=\sffamily\tiny, text=ink] at (\mx-0.10, \my+3.5*\cs) {Prompt / moves};
\node[anchor=east, font=\ttfamily\tiny, text=geostroke] at (\mx-0.10, \my+2.5*\cs) {<STEP> objs};
\node[anchor=east, font=\ttfamily\tiny, text=geostroke] at (\mx-0.10, \my+1.5*\cs) {<LOOK> objs};
\node[anchor=east, font=\sffamily\tiny, text=ink] at (\mx-0.10, \my+0.5*\cs) {Question / answer};

\node[font=\sffamily\tiny, text=muted] at (\mx+0.5*\cs, \my+4.22*\cs) {Text};
\node[font=\ttfamily\tiny, text=geostroke] at (\mx+1.5*\cs, \my+4.22*\cs) {<STEP>};
\node[font=\ttfamily\tiny, text=geostroke] at (\mx+2.5*\cs, \my+4.22*\cs) {<LOOK>};
\node[font=\sffamily\tiny, text=muted] at (\mx+3.5*\cs, \my+4.22*\cs) {Ans};
\node[font=\sffamily\tiny, text=muted] at (\mx+2.0*\cs, \my+4.58*\cs) {can attend to $\rightarrow$};

\node[font=\sffamily\tiny, text=muted] at (\mx+2.0*\cs, 0.44)
    {colored = visible \quad gray = masked};

\end{tikzpicture}}
  \caption{\textbf{Parallel spatial stream and attention mask on Qwen3-1.7B~\cite{yang2025qwen3}.} \textbf{(a)} Across each of the $28$ pretrained layers, a $256$-dimensional parallel spatial block runs alongside the $2048$-dimensional language stream ($+4.4\%$ added parameters). Language features are down-projected from $2048$ to $256$ dimensions so language and geometry enter each spatial block at equal width, and spatial updates are projected back into upper language layers ($\ell \ge 6$) through zero-initialized cross-projections ($B = 0$). \textbf{(b)} Attention visibility across the sequence: text tokens attend causally, object tokens within the same \texttt{<STEP>} or \texttt{<LOOK>} block attend to each other bidirectionally, and \texttt{<LOOK>} blocks all earlier text and intermediate steps.}
  \label{fig:zoom}
\end{figure}
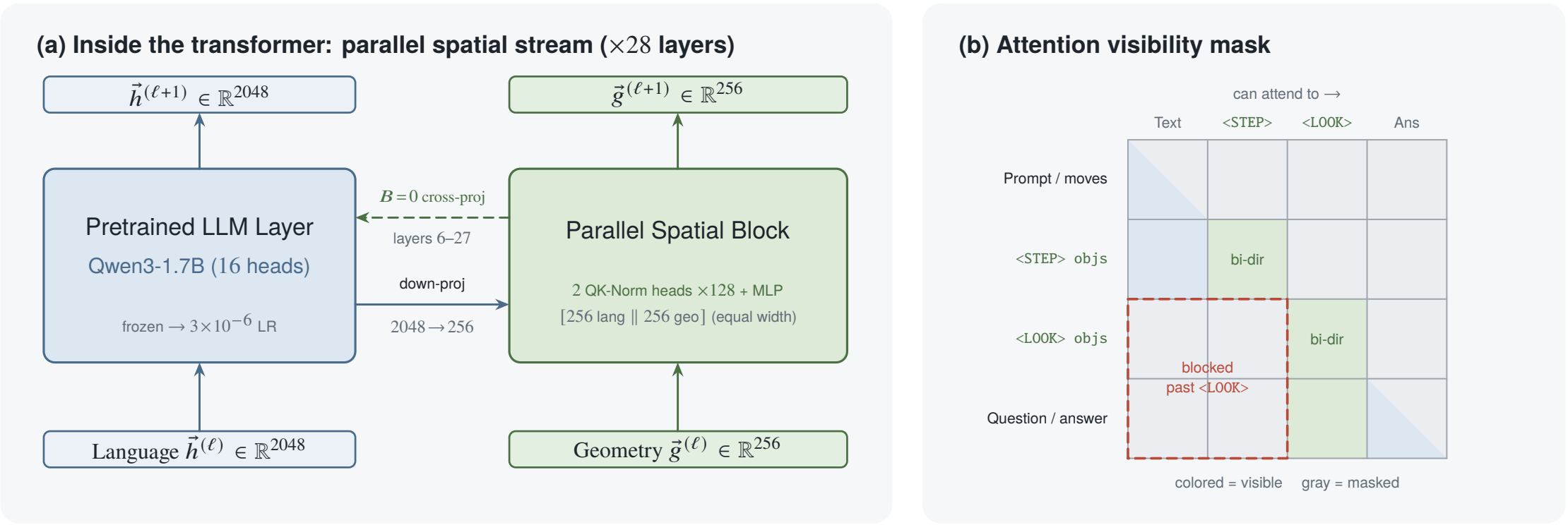

To guarantee that answers come from the current object workspace rather than earlier prose, emitting the \texttt{<LOOK>} token introduces an attention boundary (Figure~\ref{fig:zoom}b). Let $i_{\text{look}}$ denote the sequence index of the \texttt{<LOOK>} token, and let $g_i$ denote the step index of object token $i$ (shared by all objects in the same \texttt{<STEP>} or \texttt{<LOOK>} update of the object workspace, and undefined on word tokens). Sequence index $i$ is permitted to attend to sequence index $j$ if and only if
\begin{equation}
  \big(j \le i \;\lor\; g_i = g_j\big) \;\land\; \neg\big(i \ge i_{\text{look}} \;\land\; j < i_{\text{look}}\big).
  \label{eq:mask}
\end{equation}
The first clause is the standard causal attention on text and bidirectional attention on objects; the second clause blocks every token at or after \texttt{<LOOK>} from seeing anything before \texttt{<LOOK>}. Because neither the initial prompt nor the intermediate object pose updates remain visible past \texttt{<LOOK>}, the model can answer only based on the current (final) object workspace state, read back as object tokens. This makes the object workspace, geometry representation, and geometry operations testable by overriding the object representation (pose, shape) at \texttt{<LOOK>} (for example, hard-coding the objects to zero or random values) and checking that natural-language inference fails to answer the original prompt correctly.

% ============================================================================
\section{Adapting a Pretrained Language Model}
\label{sec:grafting}
% ============================================================================

\subsection{A Minimal Parallel Spatial Stream}
To implement the spatial workspace on a pretrained language model (\texttt{Qwen3-1.7B}~\cite{yang2025qwen3}, $28$ layers of width $d = 2048$) while preserving its pretrained language representations, we attach a parallel spatial stream alongside the transformer layers (Figure~\ref{fig:zoom}). Related mixed discrete-continuous sequences and modality-parallel streams have been used to adapt language models for pixel-level image diffusion~\cite{zhou2025transfusion,shi2024lmfusion}; here we apply a narrow parallel stream to non-visual object poses and signed-distance boundaries. We present this architecture as a minimal implementation for adapting an existing LLM.

For all $28$ layers (indexed by $\ell = 0, \dots, 27$), every token in the sequence (word, control, or object) is represented by two parallel state vectors: the backbone's language state $\vec{h}^{(\ell)} \in \mathbb{R}^{2048}$ ($1.72$B base parameters) and a geometry state $\vec{g}^{(\ell)} \in \mathbb{R}^{256}$ (adding $75.9$M parameters across the parallel spatial stream, projections, and output heads, or $+4.4\%$ of \texttt{Qwen3-1.7B}, comparable in parameter budget to parameter-efficient adapters~\cite{hu2022lora}). Giving word tokens their own $\vec{g}^{(\ell)}$ vector lets words hold and compare geometric quantities across layers without overwriting their natural-language meanings in $\vec{h}^{(\ell)}$.

At layer $\ell$, the pretrained language block first updates $\vec{h}^{(\ell)} \mapsto \vec{h}_{\text{lm}}^{(\ell)}$. Two projections then connect the two streams around a spatial block ($2$ normalized rotary attention heads mapping the $512$-dimensional concatenated input to $2 \times 128 = 256$ dimensions and back to $256$, followed by a $512 \rightarrow 512 \rightarrow 256$ SiLU MLP): a \emph{down-projection} $S^{(\ell)} \in \mathbb{R}^{256 \times 2048}$ maps the $2048$-element language vector down to $256$ dimensions entering the spatial block, and (for upper layers $\ell \ge 6$, with $B^{(\ell)} = 0$ fixed for $\ell < 6$) a zero-initialized \emph{cross-projection} $B^{(\ell)} \in \mathbb{R}^{2048 \times 256}$ maps the spatial block's $256$-element update $\Delta\vec{g}^{(\ell)}$ back up to $2048$ dimensions to update the language stream:
\begin{align}
  \Delta\vec{g}^{(\ell)} &= \mathrm{SpatialBlock}^{(\ell)}\!\Big(\big[\,S^{(\ell)}\,\mathrm{Norm}(\vec{h}_{\text{lm}}^{(\ell)}) \;\big\|\; \mathrm{Norm}(\vec{g}^{(\ell)})\,\big]\Big), \nonumber \\
  \vec{g}^{(\ell+1)} &= \vec{g}^{(\ell)} + \Delta\vec{g}^{(\ell)}, \qquad
  \vec{h}^{(\ell+1)} = \vec{h}_{\text{lm}}^{(\ell)} + B^{(\ell)}\,\Delta\vec{g}^{(\ell)} \quad (B^{(\ell)} = 0 \text{ for } \ell < 6).
  \label{eq:sidecar}
\end{align}
(In code, separate zero-initialized cross-projections are applied to the spatial block's attention and MLP outputs directly; we write them as a single $B^{(\ell)}\,\Delta\vec{g}^{(\ell)}$ term here for readability.) Both projections control the signal-to-noise balance. Down-projecting language from $2048$ to $256$ dimensions via $S^{(\ell)}$ ensures natural language and geometry enter every spatial attention head and MLP at equal width rather than $8\text{:}1$. Initializing every cross-projection $B^{(\ell)} = 0$ (along with a final zero-initialized projection into the word head) guarantees that at step $0$ the $B^{(\ell)}\,\Delta\vec{g}^{(\ell)}$ term vanishes, so plain-text forward passes match original \texttt{Qwen3-1.7B} before training gradually opens the connection where needed.

\subsection{Shape Generalization Across Held-Out Nouns}
To let the object head assign accurate silhouettes to everyday English nouns without requiring thousands of custom meshes, we built a reference shape table from $2{,}840$ hand-labeled noun-to-form pairs. These pairs map a broad, topic-diverse English vocabulary (spanning tools, food, furniture, clothing, and household items) onto ${\sim}50$ basic two-dimensional silhouettes (discs, rods, crescents, cups, mushroom domes, rings, triangles, L-brackets, and similar primitives). With a broad vocabulary covering those ${\sim}50$ forms and the parallel spatial stream free to learn how words map to shapes, held-out nouns omitted from the table generalize directly from pretrained word representations.

Specifically, while object poses ($\Delta\vec{p}$) and subsequent moves are predicted directly by the object head as in Section~\ref{sec:paradigm}, a newly created object's shape embedding ($\Delta\vec{z}_{32}$) is produced by matching its noun against this $2{,}840$-entry reference shape table. Each table entry stores two vectors: the noun's fixed $32$-element shape embedding $\vec{z}_{32}$ (computed once by the frozen boundary autoencoder from its basic silhouette) and a $64$-dimensional text key (computed by passing that noun through the coupled upper layers $\ell \ge 8$, combining those layers via learned softmax weights, and projecting to $64$ dimensions). When spawning a new object from words (for example, \emph{red croissant}), the model encodes that noun phrase with the exact same upper layers and $64$-dimensional projection, computes cosine similarity against all $2{,}840$ stored text keys, and outputs a softmax-weighted blend of their stored $32$-element shape embeddings $\vec{z}_{32}$, plus a bounded residual adjustment $0.15\,\tanh(\cdot)$ from the object head. Because the model's upper-layer representations evolve during training, all $2{,}840$ text keys are recomputed in a separate no-gradient pass every $200$ steps. During training, a leave-self-out mask hides each noun's own table entry whenever another training noun shares the same basic shape (leaving unique shapes visible), forcing the model to match nouns by shared physical form rather than memorizing individual token IDs.

A major benefit of a pretrained language model is that it already knows what an unseen noun looks like. Across trillions of words of ordinary prose, objects with similar geometry appear alongside the same physical verbs and constraints: slender rods roll, point, or slide into narrow slots, whereas curved shells cup, arch, or enclose. Pretrained word representations therefore already encode implicit shape structure, mixed with non-geometric facts such as topic, material, or function. Our geometry training dataset contains $2{,}840$ nouns across unrelated domains; therefore, topic becomes less relevant (for example, a kitchen \emph{plate}, a metal \emph{washer}, and a sports \emph{puck} share little topically except being flat discs). We keep \texttt{Qwen3-1.7B}'s word-embedding table fixed and train its transformer layers at a $3 \times 10^{-6}$ learning rate while backpropagating shape gradients through the coupled upper layers via $S^{(\ell)}$ and $B^{(\ell)}$. This teaches the spatial stream to isolate the directions in word space that govern physical form and discard unrelated topical features, so held-out nouns never seen during spatial training map straight onto matching $32$-element shape embeddings.

At the same time, relying on a library of roughly $50$ basic shapes has a clear limitation. New objects are formed by blending stored shapes with only a bounded residual adjustment: held-out nouns whose geometry resembles those stored shapes transfer directly, whereas nouns requiring an altogether unrepresented shape (such as a heart or a spoon when neither shape is in the library) cannot be invented from scratch by table lookup alone. Moving past this limitation requires expanding the shape library, replacing lookup with a generative shape decoder, or composing complex objects out of multiple object tokens (Section~\ref{sec:discussion}).

% ============================================================================
\section{Training Setup, Dataset Controls, and Evaluation Ablations}
\label{sec:training}
% ============================================================================

To make our reference implementation possible to reproduce or extend, we detail all data generation rules, numerical stabilization choices, optimization stages, and evaluation ablations in this section. Quantitative results follow in Section~\ref{sec:results}.

\subsection{Pretraining the Frozen Boundary Autoencoder}
Before training the language model, we pretrain the $32$-dimensional shape autoencoder (\textsf{Encoder} in Figure~\ref{fig:overview}) on synthetic two-dimensional outlines and freeze it permanently. Each training shape is centered at the origin, scaled to unit maximum radius, and sampled at $128$ evenly spaced perimeter points. The encoder applies four one-dimensional convolution layers with wrap-around padding (channels $2 \rightarrow 64 \rightarrow 128 \rightarrow 256 \rightarrow 256$, kernel width $5$, GELU activations), concatenates global max and mean pooling over the final $256$ channels across the $128$ points ($2 \times 256 = 512$ features), and maps to $\vec{z}_{32} \in \mathbb{R}^{32}$ via a two-layer MLP. The paired signed-distance decoder maps query coordinates $\vec{q} \in \mathbb{R}^2$ (expanded with six sinusoidal frequency scales) and $\vec{z}_{32}$ through a four-layer, $256$-wide MLP with a mid-network skip projection from $\vec{z}_{32}$ to scalar signed distance $\hat{d}$.

We generate $36{,}000$ outlines ($65\%$ drawn from parametric shape families and $35\%$ random polygons and smooth closed loops so corners and concavities are learned generally) and train for $9{,}000$ steps at batch size $64$ using AdamW (peak learning rate $10^{-3}$, one-cycle schedule). For each shape, we sample $192$ near-boundary points ($\sigma = 0.05$ Gaussian offset from perimeter vertices) and $192$ uniform points across $[-1.25, 1.25]^2$, and inject Gaussian noise ($\sigma = 0.03$) into $\vec{z}_{32}$ before decoding to keep the latent space smooth. We minimize the surface-clamped $L_1$ loss $|\mathrm{clamp}(\hat{d}, \pm 0.25) - \mathrm{clamp}(d, \pm 0.25)|$ plus $10^{-3}\|\vec{z}_{32}\|_2^2$. Each of the $32$ shape-embedding dimensions is then standardized to zero mean and unit variance across the training pool.

\subsection{Synthetic Scene Generation and Dataset Controls}
Training and evaluation problems are generated synthetically in $[-1, 1]^2$ by working backward from the final scene. We first place $2\text{--}5$ non-overlapping shapes and compute exact answers from their $128$-point polygons. We then apply $0\text{--}5$ reverse translations ($\Delta r \in [0.05, 0.60]$, withholding $[0.22, 0.38]$ during training) and rotations ($\Delta\theta \in [5^\circ, 175^\circ]$, withholding $[30^\circ, 60^\circ]$ during training) to obtain starting poses so that executing the forward natural-language move instructions lands on a valid final scene. Initial scenes arrive either as shown object tokens after a \texttt{<SCENE>} token (with object names omitted on a subset of these scenes) or purely as natural-language descriptions spawned via \texttt{<NEW>} tokens.

Two dataset rules prevent language models from exploiting wording cues instead of geometry:
\begin{enumerate}[leftmargin=1.5em, itemsep=2pt, topsep=2pt]
  \item \emph{Pre-scene threshold sampling:} Whenever a binary question asks about a continuous quantity (such as inter-part gap or first-contact angle, e.g., \emph{``Is the gap more than $0.35$?''}), the numerical threshold is sampled \emph{first}. Scenes or object pairs are then filtered (by rejection sampling or inverse-CDF thinning, excluding borderline values within $0.04\text{--}0.05$ distance units or $8^\circ$ of the threshold) so \texttt{yes} and \texttt{no} answers remain balanced across thresholds rather than correlating the threshold value with the answer.
  \item \emph{Word-boundary vocabulary separation:} Starting from $3{,}317$ everyday noun entries ($2{,}982$ candidate training nouns plus $332$ held-out test nouns and $3$ held-out geometric families omitted from training), any candidate training phrase containing a held-out noun as a whole word (such as \emph{mobile phone} for held-out \emph{phone}) is removed automatically ($142$ phrases dropped, leaving $2{,}840$ clean training nouns).
\end{enumerate}

\subsection{Sequence Packing and Backpropagation}
Although the model executes step by step at inference (writing its predicted objects into the object workspace and reading them back on the next step), during training each multi-step problem is assembled into a single sequence and trained in parallel across all tokens using the attention mask of Equation~\eqref{eq:mask}. On shown scenes, the prompt opens with \texttt{<SCENE>} followed by exact initial object tokens $(\vec{p}^{(0)}, \vec{z}_{32}^*)$; on prose-described scenes, the prompt text is followed by \texttt{<NEW>} declarations and an initial \texttt{<STEP>} block whose input object tokens hold zeros $(\vec{0}, \vec{0})$ and whose target updates are the full starting states $(\vec{p}^{(0)}, \vec{z}_{32}^*)$. Each subsequent move $k$ appends the move instruction, \texttt{<STEP>}, the pre-move object tokens $(\vec{p}^{(k-1)} + \vec{\epsilon}, \vec{z}_{32}^*)$, \texttt{</STEP>}, and the updated object tokens $(\vec{p}^{(k)}, \vec{z}_{32}^*)$. On move steps, we add small Gaussian jitter $\vec{\epsilon} \sim \mathcal{N}(\vec{0}, 0.01^2 I)$ to input $(x, y)$ coordinates and $-\vec{\epsilon}$ to the target update $\Delta\vec{p}^*$ so multi-step execution learns to correct small position errors. Each object token also carries a learned binary flag distinguishing pre-update inputs (inside \texttt{<STEP>}, where the object head predicts an update) from updated states (after \texttt{</STEP>} or \texttt{<LOOK>}, which are read-only).

Finally, the sequence appends \texttt{<LOOK>} followed by the final object tokens. On half of training examples, the question and target answer follow those object tokens directly; on the other half, a short natural-language scene description is inserted between the final object tokens and the question as an auxiliary training signal. In the word cross-entropy loss $\mathcal{L}_{\text{word}}$, only model-generated tokens are supervised: answer tokens are weighted $5\times$ and description tokens $3\times$ relative to control and move-step tokens ($1\times$). Because the final answer is only a few tokens long, this up-weighting concentrates gradient signal on reading the \texttt{<LOOK>} workspace accurately rather than predicting routine control formatting.

For predicted updates $(\Delta\vec{p}, \Delta\vec{z}_{32})$ and target updates $(\Delta\vec{p}^*, \Delta\vec{z}_{32}^*)$ on \texttt{<STEP>} object tokens, we minimize a joint language and object loss that weights the $32$-element shape-embedding error by $\tfrac{1}{2}$ relative to the $5$-element pose error:
\begin{equation}
  \mathcal{L}_{\text{spatial}} = \mathcal{L}_{\text{word}} + \lambda_{\text{obj}}\Big(\|\Delta\vec{p} - \Delta\vec{p}^*\|_2^2 + \tfrac{1}{2}\|\Delta\vec{z}_{32} - \Delta\vec{z}_{32}^*\|_2^2\Big), \qquad \lambda_{\text{obj}} = 8.
  \label{eq:loss}
\end{equation}
Because exact post-step object tokens are supplied in the sequence after every \texttt{</STEP>} and \texttt{<LOOK>}, one training pass updates all three parts of the workspace loop simultaneously:
\begin{enumerate}[leftmargin=1.5em, itemsep=2pt, topsep=2pt]
  \item \emph{Reading path (post-\texttt{<LOOK>}):} Because the \texttt{<LOOK>} mask blocks all prior text, $\mathcal{L}_{\text{word}}$ on answer and description tokens must be satisfied through the \texttt{<LOOK>} object tokens. Gradients pass through the word head (including the zero-initialized spatial-to-word projection), through the coupled \texttt{Qwen3-1.7B} and parallel spatial blocks, and into $W_{\text{in}}$, while the boundary autoencoder stays frozen.
  \item \emph{Pose-update path (pre-\texttt{<LOOK>}):} On \texttt{<STEP>} object tokens, the model attends to the prompt, move instructions, and prior objects. The object head reads the top-layer language state, the geometry state, and a learned weighted mix of all $28$ layers. The squared-error update loss backpropagates through this head, through the coupled attention blocks linking move instructions to objects, and into $W_{\text{in}}$.
  \item \emph{Shape-creation path:} On newly declared objects, the $\Delta\vec{z}_{32}$ loss backpropagates through the softmax matching weights computed from cosine similarities against the $2{,}840$ stored text keys. Gradients flow into the $64$-dimensional text projection, upper coupled layers ($\ell \ge 8$), and unfrozen \texttt{Qwen3-1.7B} layers as they read each noun phrase, teaching the model which semantic features predict physical shape. The stored text keys themselves are refreshed without gradients every $200$ steps.
\end{enumerate}

\subsection{Signal-to-Noise Calibration and Curriculum Schedule}
Four numerical initializations are critical when attaching new geometric layers to a pretrained transformer. First, $W_{\text{in}}$ ($\sigma = 0.25$) and the $16$-entry per-scene object-ID table ($\sigma = 0.40$ in $\mathbb{R}^{256}$, randomly assigned per scene to distinguish otherwise identical objects) are scaled so pose and shape features carry ${\sim}95\%$ of initial geometry-stream variance while object IDs carry ${\sim}5\%$; using PyTorch's default $\mathcal{N}(0, 1)$ embedding initialization gives the random object IDs roughly four times the variance of the pose and shape features, obscuring the geometry at the start of training. Second, initial geometry vectors for word and control tokens are drawn from $\mathcal{N}(0, 0.02^2)$. Third, inside every parallel spatial block, queries and keys receive per-head RMS normalization before rotary position encoding. Fourth, the object head's output layer is initialized to zero so initial state updates begin smoothly from zero.

Training proceeds in three stages summarized in Table~\ref{tab:curriculum}. Stage~0 trains and freezes the boundary autoencoder. Stage~1 runs for $4{,}000$ steps on zero-move shown scenes, training only the newly added parallel spatial stream and output heads while keeping \texttt{Qwen3-1.7B} frozen until coordinate and shape predictions are well above chance. Stage~2 runs for $16{,}000$ steps, linearly ramping task difficulty via $t = \min(1, \text{step}/9{,}600)$ over the first $60\%$ of steps and holding full difficulty through step $16{,}000$. During the first $4{,}000$ steps of Stage~2 (Stage~2a), only the parallel spatial stream and output heads train while \texttt{Qwen3-1.7B} stays frozen. At step $4{,}000$ (Stage~2b), \texttt{Qwen3-1.7B}'s $28$ transformer layers unfreeze at a learning rate of $3 \times 10^{-6}$ so language and spatial layers refine together.

During Stages~1 and~2, $80\%$ of update steps optimize batches of $8$ spatial examples (padded to the longest sequence in the batch), and $20\%$ optimize batches of $8 \times 256$-token English-text passages from WikiText-2 using next-token cross-entropy plus $\mathrm{KL}(p_{\text{base}} \,\|\, p_{\theta})$ against original \texttt{Qwen3-1.7B}. New weights train with AdamW at learning rate $2 \times 10^{-4}$ (after a $300$-step linear warmup) and weight decay $0.01$; unfrozen \texttt{Qwen3-1.7B} layers use weight decay $0.0$, global gradient norm clipping at $1.0$, and a fixed input word-embedding table throughout. On a single RTX PRO 6000 GPU, Stage~0 takes under two minutes and Stages~1 and~2 together take ${\sim}3.8$ hours.

\begin{table}[!ht]
  \centering
  \footnotesize
  \caption{\textbf{Training stage order, batch composition, and linear curriculum schedule.} Stage~2 linearly interpolates $t = \min(1, \text{step}/9{,}600)$ across Stages~2a and~2b from step $0$ to $9{,}600$ (ramping question weights from $(5, 2, 0.5, 0.2, 1)$ to $(2, 1.5, 2.5, 1.5, 1)$) and holds the $t=1$ distribution through step $16{,}000$. Question weights give relative sampling shares across the five basic question types (\texttt{relation}, \texttt{closest}, \texttt{touch}, \texttt{gap}, \texttt{name}); \emph{scenarios} are multi-constraint spatial problems.}
  \label{tab:curriculum}
  \vspace{3pt}
  \begin{tabular}{@{}>{\raggedright\arraybackslash}p{0.18\textwidth}>{\raggedright\arraybackslash}p{0.20\textwidth}>{\raggedright\arraybackslash}p{0.235\textwidth}>{\raggedright\arraybackslash}p{0.30\textwidth}@{}}
    \toprule
    \textbf{Stage \& Steps} & \textbf{Batch Structure} & \textbf{Trainable Weights \& LR} & \textbf{Task \& Curriculum Mix} \\
    \midrule
    Stage 0: Autoencoder \newline ($9{,}000$ steps) & $64$ outlines / step \newline ($384$ signed-distance \newline points / outline) & Encoder + signed-distance \newline decoder (peak LR $10^{-3}$; \newline frozen afterward) & $65\%$ parametric shapes, $35\%$ random loops/polygons; $\sigma = 0.03$ noise on $\vec{z}_{32}$ \\[5pt]
    Stage 1: Object Reading \newline ($4{,}000$ steps) & $80\%$: $8$ spatial examples \newline $20\%$: $8 \times 256$-token \newline WikiText-2 & Parallel stream + heads \newline (LR $2 \times 10^{-4}$) \newline \texttt{Qwen3-1.7B} frozen & $0$ moves; $100\%$ shown (\texttt{<SCENE>}); $0\%$ scenarios; question weights $(5, 2, 0, 0, 0)$ \\[5pt]
    Stage 2a: Curriculum \newline (steps $0\text{--}4{,}000$) & $80\%$: $8$ spatial examples \newline $20\%$: $8 \times 256$-token \newline WikiText-2 & Parallel stream + heads \newline (LR $2 \times 10^{-4}$) \newline \texttt{Qwen3-1.7B} frozen & $0\text{--}1 \!\rightarrow\! 0\text{--}5$ moves; prose-spawned (\texttt{<NEW>}): $15\% \!\rightarrow\! 40\%$; scenarios: $5\% \!\rightarrow\! 30\%$; yes/no: $80\% \!\rightarrow\! 40\%$ \\[5pt]
    Stage 2b: Coupled Run \newline (steps $4{,}000\text{--}16{,}000$) & $80\%$: $8$ spatial examples \newline $20\%$: $8 \times 256$-token \newline WikiText-2 & Parallel stream + heads \newline (LR $2 \times 10^{-4}$) \newline \texttt{Qwen3-1.7B} (LR $3 \times 10^{-6}$, word table fixed) & Curriculum completes at step $9{,}600$, then held constant; question weights $\rightarrow (2, 1.5, 2.5, 1.5, 1)$ \\
    \bottomrule
  \end{tabular}
\end{table}

\subsection{Workspace Ablation Protocol for Evaluation}
Even with careful dataset construction, claiming that a language model reasons through a continuous workspace requires direct ablation checks at evaluation time. After training, we test the model on identical held-out problems across four conditions applied to the object tokens inside the \texttt{<LOOK>} block (both when answering directly from object tokens and when generating a scene description first):
\begin{enumerate}[leftmargin=1.5em, itemsep=2pt, topsep=2pt]
  \item \emph{Normal:} the model generates its own \texttt{<NEW>} and \texttt{<STEP>} updates and reads its own resulting workspace at \texttt{<LOOK>};
  \item \emph{Exact \texttt{<LOOK>}:} intermediate \texttt{<STEP>} updates remain model-generated, and only the final \texttt{<LOOK>} object tokens are replaced by exact object states, isolating reading error from multi-step writing drift;
  \item \emph{Zeroed \texttt{<LOOK>}:} all object tokens inside \texttt{<LOOK>} are replaced by zeros, testing what the language model can infer when the workspace state is withheld; and
  \item \emph{Randomized \texttt{<LOOK>}:} final \texttt{<LOOK>} positions and shape embeddings are replaced with new random draws ($x, y \sim \mathcal{U}[-0.8, 0.8]$, $\vec{z}_{32} \sim \mathcal{N}(\vec{0}, I)$).
\end{enumerate}
True workspace reliance requires \emph{Normal} accuracy to track closely with \emph{Exact}, while collapsing toward chance under both \emph{Zeroed} and \emph{Randomized} ablations.

% ============================================================================
\section{Experiments and Results}
\label{sec:results}
% ============================================================================

We evaluate the adapted \texttt{Qwen3-1.7B} model (with no external geometry tools at inference) across three core capabilities: (1) executing multi-step spatial manipulation from natural-language descriptions, (2) generalizing shapes to held-out nouns omitted from the reference shape table (training data), and (3) reading final answers causally from the object workspace while preserving ordinary English fluency. Unless otherwise noted, all evaluations in this section generate answers directly from the object tokens after \texttt{<LOOK>}, with no intermediate text description of the scene.

\subsection{Multi-Step Spatial Manipulation from Prose}
The workspace states are decoded from the model's $32$-element shape embeddings and $5$-element poses, pairing each execution with its prompt and head-switching sequence. Each \texttt{<NEW>} declaration ends with the fixed cue ``\texttt{(shape:}'' to trigger shape matching against the reference table. Each \texttt{<STEP>} applies a discrete pose update $\Delta\vec{p}$ without intermediate collision checking, so objects can pass freely through each other between steps.

\begin{figure}[H]
  \vspace{-4pt}
  \centering
  \promptcard{%
    \textbf{Prompt:} \texttt{Problem: - a blue cup, size 0.34, at x 0.15, y -0.15, turned 0 degrees / - a red coin, size 0.06, at x -0.45, y 0.35, turned 0 degrees. Steps: 1. Move the red coin 0.60 to the right. 2. Move the red coin 0.50 down. Question: Does the red coin sit within the hollow of the blue cup? Solution:}}{%
    \textbf{Sequence} (\whead{word head}\;\; \ohead{\text{object head}}\;\; \wsys{workspace objects}):\;
    \whead{<NEW> blue cup (shape:}\; \whead{<NEW> red coin (shape:}\; \whead{<STEP>}\; \ohead{\{(\Delta\vec{p}_k,\Delta\vec{z}_k)\}_{k=1}^2}\; \wsys{</STEP>[o1,o2]}\\[1.5pt]
    \whead{Move the red coin 0.60 to the right. <STEP>}\; \ohead{\Delta\vec{p}_{1..2}}\; \wsys{</STEP>[o1,o2]}\;
    \whead{Move the red coin 0.50 down. <STEP>}\; \ohead{\Delta\vec{p}_{1..2}}\; \wsys{</STEP>[o1,o2]}\\[1.5pt]
    \whead{<LOOK>}\; \wsys{[o1,o2] Question: Does the red coin sit within the hollow of the blue cup? Answer:}\; \whead{yes}}
  \vspace{1pt}
  \includegraphics[width=0.76\textwidth]{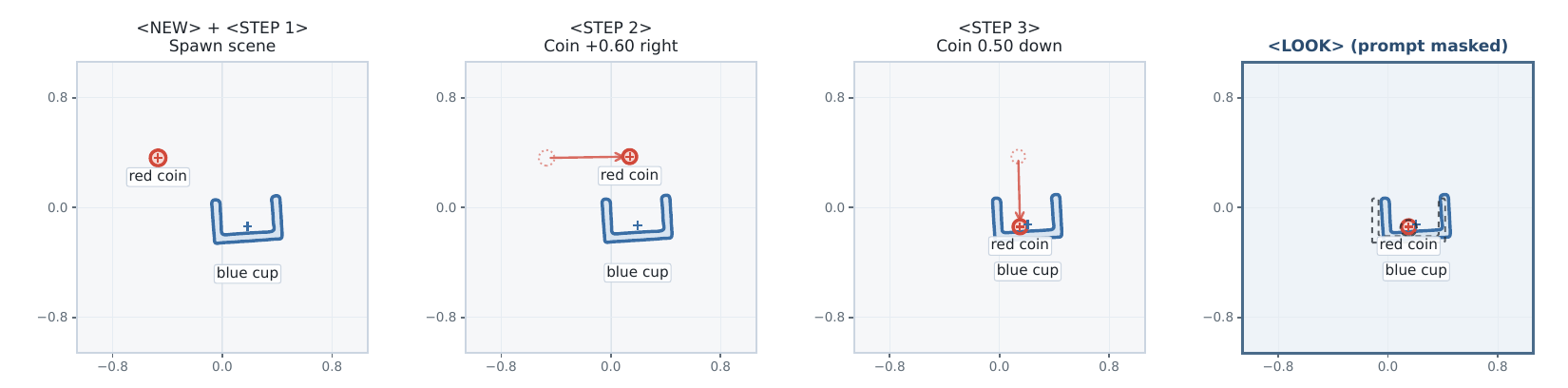}
  \vspace{-4pt}
  \caption{\textbf{Judging whether one object sits inside another after step updates.} Colored contours decode $32$-element shape embeddings and $5$-element poses; dark dashed outlines on \texttt{<LOOK>} show exact target polygons.}
  \label{fig:cup_coin}
  \vspace{-6pt}
\end{figure}

First, we demonstrate the model's capability with two objects and two translations. The model spawns a \texttt{blue cup} (part of the training dataset) and a \texttt{red coin} (not in the training dataset), applies two instructed coordinate updates placing the coin inside the cup's interior, and reads \texttt{<LOOK>} to answer ``\texttt{yes}''. The user prompt, LLM inference, and workspace object state (post-processed with external graphics software) are shown in Figure~\ref{fig:cup_coin}.

\begin{figure}[H]
  \vspace{-4pt}
  \centering
  \promptcard{%
    \textbf{Prompt:} \texttt{Problem: - a blue crate, size 0.24, at x -0.55, y -0.25, turned 0 degrees / - a red pyramid, size 0.24, at x 0.55, y 0.20, turned 90 degrees / - a yellow coin, size 0.10, at x 0.00, y 0.62, turned 0 degrees. Steps: 1. Move the blue crate 0.55 to the right. 2. Turn the red pyramid 90 degrees counter-clockwise. 3. Move the red pyramid 0.55 to the left. Question: What is closest to the yellow coin? Solution:}}{%
    \textbf{Sequence:}\;
    \whead{<NEW> blue crate (shape:}\; \whead{<NEW> red pyramid (shape:}\; \whead{<NEW> yellow coin (shape:}\; \whead{<STEP>}\; \ohead{\{(\Delta\vec{p}_k,\Delta\vec{z}_k)\}_{k=1}^3}\; \wsys{</STEP>[o1..o3]}\\[1.5pt]
    \whead{Move the blue crate 0.55 to the right. <STEP>}\; \ohead{\Delta\vec{p}_{1..3}}\; \wsys{</STEP>[o1..o3]}\;
    \whead{Turn the red pyramid 90 degrees counter-clockwise. <STEP>}\; \ohead{\Delta\vec{p}_{1..3}}\; \wsys{</STEP>[o1..o3]}\\[1.5pt]
    \whead{Move the red pyramid 0.55 to the left. <STEP>}\; \ohead{\Delta\vec{p}_{1..3}}\; \wsys{</STEP>[o1..o3]}\;
    \whead{<LOOK>}\; \wsys{[o1..o3] Question: What is closest to the yellow coin? Answer:}\; \whead{red pyramid}}
  \vspace{1pt}
  \includegraphics[width=0.84\textwidth]{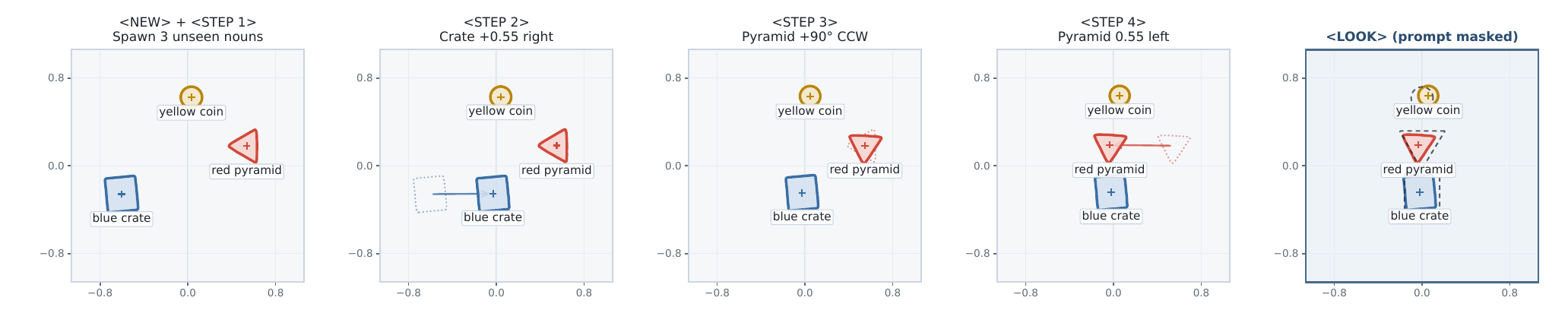}
  \vspace{-4pt}
  \caption{\textbf{Assembling scattered held-out nouns across translations and rotation.} Dotted outlines and arrows mark the previous step; dark dashed outlines on \texttt{<LOOK>} show exact target polygons.}
  \label{fig:house}
  \vspace{-4pt}
\end{figure}

Next, we test three nouns (\texttt{blue crate}, \texttt{red pyramid}, and \texttt{yellow coin}) held out from the training dataset. Starting from the initial object positions with one of them (\texttt{pyramid}) initialized on its side (rotated $90^\circ$), the model shifts the crate to the center, rotates the pyramid $90^\circ$ counter-clockwise (to $180^\circ$), shifts it atop the crate, and reads \texttt{<LOOK>} to answer ``\texttt{red pyramid}''. The user prompt, LLM inference, and workspace object state (post-processed with external graphics software) are shown in Figure~\ref{fig:house}.

\begin{figure}[H]
  \vspace{-4pt}
  \centering
  \promptcard{%
    \textbf{Prompt:} \texttt{Problem: - a red crayon, size 0.30, at x 0.10, y 0.05, turned 90 degrees / - a green coin, size 0.08, at x 0.10, y -0.42, turned 0 degrees / - a yellow umbrella, size 0.17, at x 0.15, y 0.50, turned 0 degrees / - a blue phone, size 0.16, at x -0.50, y 0.48, turned 0 degrees. Steps: 1. Turn the red crayon 90 degrees clockwise. 2. Move the red crayon 0.55 to the left. 3. Move the green coin 0.55 up. Question: What is closest to the yellow umbrella? Solution:}}{%
    \textbf{Sequence} (\whead{word head}\;\; \ohead{\text{object head}}\;\; \wsys{workspace objects}):\;
    \whead{<NEW> red crayon ... <NEW> blue phone (shape:}\; \whead{<STEP>}\; \ohead{\{(\Delta\vec{p}_k,\Delta\vec{z}_k)\}_{k=1}^4}\; \wsys{</STEP>[o1..o4]}\\[1.5pt]
    \whead{Turn the red crayon 90 degrees clockwise. <STEP>}\; \ohead{\Delta\vec{p}_{1..4}}\; \wsys{</STEP>[o1..o4]}\;
    \whead{Move the red crayon 0.55 to the left. <STEP>}\; \ohead{\Delta\vec{p}_{1..4}}\; \wsys{</STEP>[o1..o4]}\\[1.5pt]
    \whead{Move the green coin 0.55 up. <STEP>}\; \ohead{\Delta\vec{p}_{1..4}}\; \wsys{</STEP>[o1..o4]}\;
    \whead{<LOOK>}\; \wsys{[o1..o4] Question: What is closest to the yellow umbrella? Answer:}\; \whead{green coin}}
  \vspace{1pt}
  \includegraphics[width=0.84\textwidth]{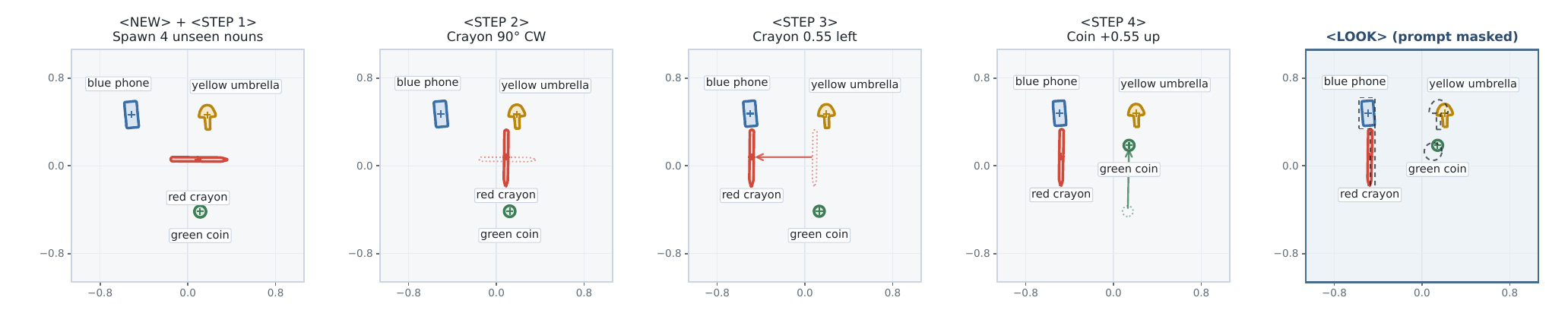}
  \vspace{-4pt}
  \caption{\textbf{Multi-step rotation and nearest-object question.} Turning a horizontal \texttt{red crayon} $90^\circ$ upright, shifting it left, and moving the \texttt{green coin} beneath the \texttt{yellow umbrella}.}
  \label{fig:unjam}
  \vspace{-6pt}
\end{figure}

Next, we test multi-step rotation and translation across four held-out nouns (\texttt{red crayon}, \texttt{green coin}, \texttt{yellow umbrella}, and \texttt{blue phone}), starting with the \texttt{red crayon} lying horizontally ($90^\circ$) between the \texttt{green coin} and \texttt{yellow umbrella}. Across three instructed moves, the model rotates the crayon $90^\circ$ clockwise upright, shifts it to the left, moves the \texttt{green coin} up beneath the \texttt{yellow umbrella}, and reads \texttt{<LOOK>} to answer ``\texttt{green coin}''. The user prompt, LLM inference, and workspace object state (post-processed with external graphics software) are shown in Figure~\ref{fig:unjam}.

\begin{figure}[H]
  \vspace{-4pt}
  \centering
  \promptcard{%
    \textbf{Prompt:} \texttt{Problem: There is a red croissant. A blue umbrella is directly above the red croissant. A green phone is directly to the left of the blue umbrella. A yellow starfish is directly below the green phone. Question: Is the yellow starfish to the left of the red croissant? Solution:}}{%
    \textbf{Sequence:}\;
    \whead{<NEW> red croissant (shape:}\; \whead{<NEW> blue umbrella (shape:}\; \whead{<NEW> green phone (shape:}\\[1.5pt]
    \whead{<NEW> yellow starfish (shape:}\; \whead{<STEP>}\; \ohead{\{(\Delta\vec{p}_k,\Delta\vec{z}_k)\}_{k=1}^4}\; \wsys{</STEP>[o1..o4]}\\[1.5pt]
    \whead{<LOOK>}\; \wsys{[o1..o4] Question: Is the yellow starfish to the left of the red croissant? Answer:}\; \whead{yes}}
  \vspace{1pt}
  \includegraphics[width=0.44\textwidth]{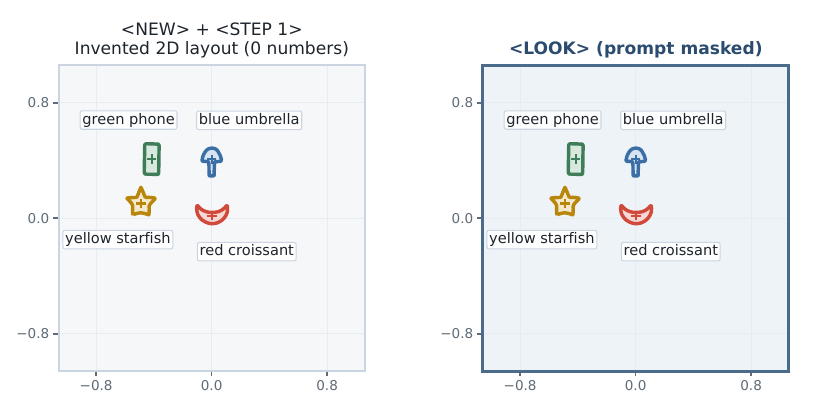}
  \vspace{-4pt}
  \caption{\textbf{Coordinate-free scene layout.} The user prompt contains only relative natural-language statements with no coordinates or sizes; the model constructs a $2 \times 2$ workspace layout across four held-out nouns and answers from \texttt{<LOOK>}.}
  \label{fig:zero_coords}
\end{figure}

Finally, we test an input prompt with no numerical values (no coordinates, angles, or sizes), given only relative natural-language statements across four held-out nouns (\texttt{red croissant}, \texttt{blue umbrella}, \texttt{green phone}, and \texttt{yellow starfish}). The model constructs a $2 \times 2$ spatial layout on the first \texttt{<STEP>} and reads \texttt{<LOOK>} to answer an unstated transitive relation (\texttt{left of?} $\rightarrow$ ``\texttt{yes}''). The user prompt, LLM inference, and workspace object state (post-processed with external graphics software) are shown in Figure~\ref{fig:zero_coords}.

Across $200$ synthetic coordinate-free problems of $3\text{--}4$ relative placements ($100$ with trained nouns and $100$ with held-out nouns), the model produces a valid workspace sequence on every problem and satisfies all stated spatial relations on $100\%$ of trained-noun problems and $99\%$ of held-out-noun problems. On the unstated \texttt{yes}/\texttt{no} questions, accuracy is $100\%$ across all $200$ problems, but drops to random-chance levels (${\sim}50\%$) when the workspace objects after \texttt{<LOOK>} are zeroed or randomized.

Across Figures~\ref{fig:cup_coin}--\ref{fig:zero_coords}, the model creates continuous objects from text, arranges and manipulates them in the workspace as instructed, and answers each spatial question correctly by inspecting the resulting workspace.

\subsection{Shape Generalization to Held-Out Nouns}
\begin{figure}[H]
  \centering
  \includegraphics[width=\textwidth]{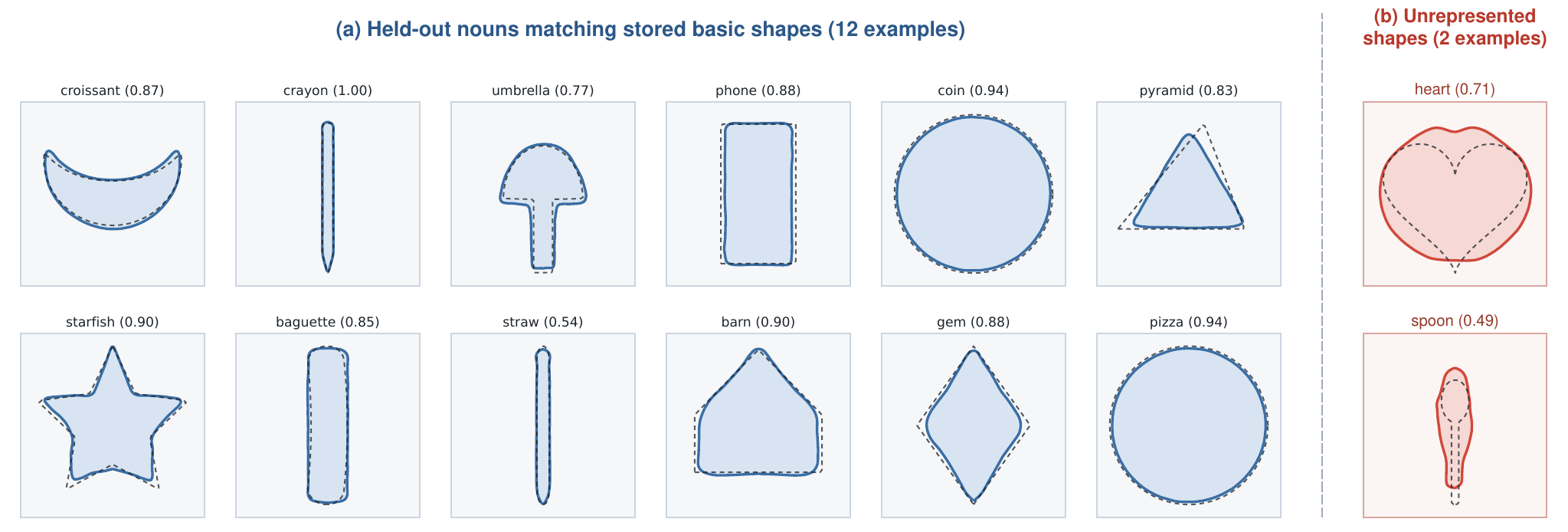}
  \caption{\textbf{Silhouettes generated for $14$ held-out nouns omitted from spatial training.} Numbers in parentheses report intersection-over-union (IoU) against the exact reference shapes (dark dashed outlines). \textbf{(a)} Twelve held-out nouns (blue) whose geometry resembles basic shapes in the ${\sim}50$-shape library. \textbf{(b)} Two held-out shapes (red: \emph{heart}, \emph{spoon}) absent from the library, illustrating the shape-embedding limitation described in Section~\ref{sec:grafting}.}
  \label{fig:gallery}
\end{figure}

Next, we evaluate the model's capability to spawn everyday nouns omitted from spatial training (all compound training phrases containing those words were removed from the training data as well; see Section~\ref{sec:training}). Figure~\ref{fig:gallery} shows $14$ representative held-out examples split into two groups. Panel~(a) displays $12$ held-out nouns whose geometry resembles shapes in the ${\sim}50$-shape reference library: curved arcs (\emph{croissant}), slender rods (\emph{crayon}, \emph{baguette}, \emph{straw}), domed canopies (\emph{umbrella}), rectangles (\emph{phone}), circular discs (\emph{coin}, \emph{pizza}), triangles and diamonds (\emph{pyramid}, \emph{gem}), five-pointed stars (\emph{starfish}), and pitched outlines (\emph{barn}). Panel~(b) displays $2$ held-out shapes (\emph{heart}, \emph{spoon}) absent from the library.

Table~\ref{tab:unseen} summarizes shape creation accuracy across trained nouns, the full set of $32$ held-out nouns whose shapes resemble entries in the ${\sim}50$-shape library, and all $3$ unrepresented shapes (\emph{heart}, \emph{spoon}, \emph{T-shape}) absent from the library. The first row reports the autoencoder given exact outlines directly ($0.98$ IoU on trained and held-out nouns), establishing an upper bound on shape reconstruction. The second row reports our adapted \texttt{Qwen3-1.7B} model generating shapes from text alone: it achieves $0.91$ IoU ($98\%$ basic-shape accuracy) on trained nouns and generalizes to $0.79$ IoU ($74\%$ basic-shape accuracy) on the $32$ held-out nouns. On the $3$ unrepresented shapes, the adapted model stays at $0.63$ IoU (versus $0.92$ for the autoencoder) because no simple combination of the stored basic shapes matches a heart, spoon, or T-shape.

\begin{table}[H]
  \centering
  \footnotesize
  \caption{\textbf{Shape creation accuracy across trained nouns, $32$ held-out nouns, and $3$ unrepresented shapes.} Shape IoU measures overlap between the decoded $32$-element shape embedding and the exact target polygon; Basic Shape measures whether the nearest basic shape matches the target.}
  \label{tab:unseen}
  \vspace{3pt}
  \begin{tabular*}{\textwidth}{@{\extracolsep{\fill}}lccccc@{}}
    \toprule
    & \multicolumn{2}{c}{\textbf{Trained Nouns}} & \multicolumn{2}{c}{\textbf{Held-Out Nouns ($32$)}} & \textbf{Unrepresented ($3$)} \\
    \cmidrule(lr){2-3} \cmidrule(lr){4-5} \cmidrule(l){6-6}
    \textbf{Model} & Shape IoU & Basic Shape & Shape IoU & Basic Shape & Shape IoU \\
    \midrule
    Autoencoder (exact outline input) & $0.98$ & --- & $0.98$ & --- & $0.92$ \\
    Adapted \texttt{Qwen3-1.7B} (text input) & $0.91$ & $98\%$ & $0.79$ & $74\%$ & $0.63$ \\
    \bottomrule
  \end{tabular*}
\end{table}

\subsection{Workspace Ablations and Language Retention}
Table~\ref{tab:sabotage} evaluates the four \texttt{<LOOK>} ablation conditions from Section~\ref{sec:training} across $300$ identical held-out problems ($0\text{--}3$ moves, $40\%$ multi-constraint scenarios). Under \emph{Direct from objects}, the model answers directly from the object tokens after \texttt{<LOOK>}; under \emph{Describe then answer}, it first writes a short text description of the workspace before answering. Columns break down overall accuracy (\emph{All}) into four question types: naming an unnamed object strictly from its shape embedding (\emph{Shape ID}), answering relative-position and nearest-object questions (\emph{Position}), judging contact and clearance distances between object outlines (\emph{Form}), and solving multi-constraint puzzles such as containment or sliding until contact (\emph{Scenarios}).

Across all $300$ problems, the model emits valid control sequences $100\%$ of the time and tracks final object positions and scales within $0.03$ mean absolute error on $[-1, 1]^2$. When answering directly from object tokens, normal workspace execution ($71\%$) matches \emph{Exact \texttt{<LOOK>}} ($70\%$), showing that multi-step workspace updates cause no loss in downstream reasoning accuracy. Under \emph{Direct from objects}, zeroing or randomizing the \texttt{<LOOK>} object tokens reduces overall accuracy to $38\%$ and $36\%$, respectively, with sharp drops in every category: \emph{Shape ID} falls from $100\%$ to $0\%$, \emph{Position} from $98\%$ to $54\text{--}57\%$, \emph{Form} from $55\%$ to $26\text{--}30\%$, and \emph{Scenarios} from $64\%$ to $39\text{--}41\%$. Lastly, writing a text description after \texttt{<LOOK>} before answering shows the same drop ($69\%$ normal versus $35\%$ and $39\%$ when zeroed or randomized, respectively), confirming that the model relies on the workspace rather than earlier text and gains no advantage from describing the scene first.

\begin{table}[H]
  \centering
  \footnotesize
  \caption{\textbf{Workspace ablations across $300$ identical held-out problems.} Across full multi-step runs, valid sequence completion is $100\%$, mean final position error is $0.03$, and scale error is $0.03$.}
  \label{tab:sabotage}
  \vspace{3pt}
  \begin{tabular*}{\textwidth}{@{\extracolsep{\fill}}llccccc@{}}
    \toprule
    \textbf{Answer Order} & \textbf{\texttt{<LOOK>} Condition} & \textbf{All} & \textbf{Shape ID} & \textbf{Position} & \textbf{Form} & \textbf{Scenarios} \\
    \midrule
    Direct from objects & Normal workspace & $71\%$ & $100\%$ & $98\%$ & $55\%$ & $64\%$ \\
    & Exact \texttt{<LOOK>} (true state) & $70\%$ & $100\%$ & $96\%$ & $58\%$ & $61\%$ \\
    & Zeroed \texttt{<LOOK>} (all zeros) & $38\%$ & $0\%$ & $54\%$ & $30\%$ & $41\%$ \\
    & Randomized \texttt{<LOOK>} (random) & $36\%$ & $0\%$ & $57\%$ & $26\%$ & $39\%$ \\
    \midrule
    Describe then answer & Normal workspace & $69\%$ & $100\%$ & $98\%$ & $53\%$ & $61\%$ \\
    & Exact \texttt{<LOOK>} (true state) & $68\%$ & $100\%$ & $98\%$ & $48\%$ & $61\%$ \\
    & Zeroed \texttt{<LOOK>} (all zeros) & $35\%$ & $0\%$ & $50\%$ & $32\%$ & $36\%$ \\
    & Randomized \texttt{<LOOK>} (random) & $39\%$ & $0\%$ & $59\%$ & $24\%$ & $46\%$ \\
    \bottomrule
  \end{tabular*}
\end{table}

Finally, Table~\ref{tab:english} shows English language retention across $128$ blocks of $256$ tokens ($128 \times 255 = 32{,}640$ next-token predictions) of the WikiText-2 test split. Lower perplexity indicates better next-word prediction. Adapting \texttt{Qwen3-1.7B} increases perplexity by $+2.34$ ($16.38 \rightarrow 18.72$): $+1.81$ comes from the adapted \texttt{Qwen3-1.7B} layers with cross-projections disabled ($18.19$), and $+0.53$ from enabling the active cross-projections. English inference remains coherent rather than suffering catastrophic forgetting ($70.2\%$ exact top-1 next-token match across the $151{,}936$-token vocabulary, $\mathrm{KL} = 0.50$ nats), and the model emits $0$ spurious workspace control tokens across all $32{,}640$ English-text positions ($0.00\%$).

\begin{table}[H]
  \vspace{-3pt}
  \centering
  \footnotesize
  \setlength{\tabcolsep}{3.5pt}
  \caption{\textbf{English language retention on WikiText-2 test ($128 \times 256$-token blocks, $32{,}640$ next-token predictions; lower PPL is better).} Full adapted model evaluation includes all parallel spatial layers, cross-projections, and control heads.}
  \label{tab:english}
  \vspace{2pt}
  \begin{tabular*}{\textwidth}{@{\extracolsep{\fill}}lcccc@{}}
    \toprule
    \textbf{Model} & \textbf{WikiText-2 PPL ($\downarrow$)} & \textbf{Top-1 Agreement} & \textbf{Mean $\mathrm{KL}(p_{\text{base}} \| p_{\theta})$} & \textbf{Control Token Rate} \\
    \midrule
    Original \texttt{Qwen3-1.7B-Base} & $16.38$ & --- & $0.00$ & --- \\
    Adapted \texttt{Qwen3-1.7B} (cross-proj.\ off) & $18.19$ & --- & --- & --- \\
    Full adapted model ($+4.4\%$ active) & $18.72$ & $70.2\%$ & $0.50$ nats & $0.00\%$ ($0 / 32{,}640$) \\
    \bottomrule
  \end{tabular*}
  \vspace{-3pt}
\end{table}

Taken together, the results in Section~\ref{sec:results} establish three empirical findings. First, the model creates, transforms, and reads continuous shapes inside one sequence without external geometry tools. Second, zeroing or randomizing \texttt{<LOOK>} significantly reduces accuracy, confirming that answers rely on the workspace rather than earlier text. Third, physical knowledge already present in pretrained language weights enables held-out everyday nouns to take shape and be manipulated consistently while keeping natural language coherent, showing that discrete language and continuous space can interact inside a single modified language model.

% ============================================================================
\section{Future Work}
\label{sec:discussion}
% ============================================================================

Because object tokens are agnostic to what continuous state they carry, two-dimensional rigid shapes represent only the entry point of a wider research program across four connected directions.

\emph{Exploring new architectures and training methods.} In this work, we attached a narrow parallel spatial stream to a pretrained language model as a minimal proof of concept. A direct architectural alternative is co-training word and object tokens from scratch inside a single shared transformer with no separate parallel stream, allowing linguistic and geometric representations to form together within the same weights. Likewise, replacing table-based shape lookup with a generative shape decoder (or composing complex objects from multiple object tokens) would allow arbitrary silhouettes beyond a fixed library, while scaling to larger models and reinforcement-learning post-training could strengthen multi-step spatial planning.

\emph{Three-dimensional geometry and hierarchical assemblies.} Extending the workspace from planar $\mathrm{SE}(2)$ to three-dimensional $\mathrm{SE}(3)$ is conceptually direct: the five-element pose becomes a ten-element frame combining $(x, y, z)$, a smooth six-element rotation representation (two orthonormal columns of $R \in \mathrm{SO}(3)$, avoiding gimbal lock and quaternion discontinuities), and scale, paired with a three-dimensional signed-distance shape embedding~\cite{park2019deepsdf} trained over solid meshes. Operating over a handful of object tokens rather than tens of thousands of image patches keeps geometric attention inexpensive: local-frame relative pose biases ($\Delta\vec{p}_{i \leftarrow j} = R_i^\top(\vec{p}_j - \vec{p}_i)/s_i$) and AlphaFold-style three-body triangle consistency~\cite{jumper2021alphafold} cost a small fraction of a single language layer. Scaling from individual parts to machines or facilities with thousands of components can follow how engineers chunk complexity through a hierarchical \texttt{<ZOOM>} operation: treating an entire pump assembly as one coarse token at room scale, then expanding that token into its internal housing, impeller, shaft, and fasteners when reasoning about internal fit.

\emph{Physics, mechanics, deformation, and material properties.} Because \texttt{<STEP>} already predicts $\Delta\vec{z}$ alongside $\Delta\vec{p}$ (with $\Delta\vec{z} = \vec{0}$ supervised only in rigid scenes), allowing non-zero shape updates while appending continuous state channels turns the workspace from kinematic motion into intuitive mechanics: object tokens can carry velocity, acceleration, mass, friction, stiffness, and ductility alongside pose and shape, so pressing a rigid wedge into a malleable bar permanently dents its shape embedding while an elastic body compresses under load and rebounds on the next step. Incorporating variable, event-driven time increments $\Delta t$ lets a model step across free motion and slow down around collisions, tipping thresholds, or mechanical jams. Following Sutton's bitter lesson~\cite{sutton2019bitter}, physical tokens should hard-code only the coordinate frame of spacetime $(x, y, z, R, \Delta t)$ alongside unconstrained continuous state vectors. Conservation, leverage, acceleration, and material response can then be learned from predicting trajectories across diverse physical environments rather than hand-crafted textbook formulas.

\emph{Robotics and two-way embodied control.} Contemporary vision-language-action models~\cite{black2024pi0} attach a motor head at the tail end of a vision-language model: information flows one way from camera pixels out to joint angles, with external objects never represented as persistent entities that can answer back to language. Replacing that one-way exit with a shared object workspace closes the loop in both directions. Camera, depth, and tactile encoders can supply the exact same continuous object tokens that language constructs from words; before moving a limb or tool, a robot can step candidate grasps, insertions, or tool paths inside its workspace, detect where a part binds or wrist torque falls short, revise its plan in language, and act out of physical foresight rather than open-loop reflex.

% ============================================================================
\section{Conclusion}
\label{sec:conclusion}
% ============================================================================

In biological evolution, space, time, force, and physical cause came hundreds of millions of years before a single word was spoken. Long before human language appeared, nervous systems already knew how solid bodies occupy volume, how rotation changes clearance, how momentum transfers across contact, and how effort meets resistance; words arrived at the very end as secondary symbols built to point back to those lived objects and forces. Artificial intelligence proceeded in the opposite order. By training large transformers on human text alone, the field mastered those secondary representations of representations first, recovering logic, mathematics, and code at a remarkable scale before ever giving models direct access to simulated objects, continuous space, or physical resistance.

That inversion carried far enough to create a powerful illusion, because centuries of human thought had already distilled vast causal structure into prose. Yet written language contains a permanent blind spot: people rarely write down the elementary geometry of the world, because every human reader already runs it silently inside the parietal cortex. An author writes that a key jammed or a beam cleared a doorway without ever stating the continuous boundary distances and angular sweeps that governed the outcome. Trained only on discrete transcripts of what humans chose to mention, a pure language model resembles a mind locked in a dark room receiving telegraph messages about a world it has never touched, forced to imitate spatial understanding through verbal ingenuity while remaining blind to form.

Closing that gap does not require flooding language models with camera pixels or rendering internal movies. As congenital aphantasia and spatial mastery in the blind reveal, the mind's grasp of geometry does not depend on a visual image behind the eyes; at its core, it is an invisible, tactile sense of coordinate frames, solid surfaces, and transformations. This work shows that it is possible to place continuous shapes and poses directly into the transformer sequence alongside words, and let language and spatial weights mingle across layers, giving a disembodied model something closer to virtual hands and mathematical touch than to artificial sight. Words no longer have to impersonate physical reality; they return to their proper role as compressed pointers that summon, manipulate, and read a continuous medium.

Two centuries ago, Schopenhauer observed that intelligence rests on four distinct roots of sufficient reason that cannot be reduced to one another: conceptual \emph{knowing} in language, geometric \emph{being} in space and time, causal \emph{becoming} across physical interaction, and goal-directed \emph{will} in action. Text transformers isolated and conquered the first root. Continuous spatial workspaces offer a direct foothold on the second, and a natural bridge toward dynamics and embodied action in the third and fourth. When discrete language and continuous physical intuition share the same sequence from the ground up, artificial intelligence need not remain an eloquent spectator standing outside the physical world looking in; it can begin to think inside the same space and time we inhabit.

\clearpage
{\small
\setlength{\bibsep}{3.5pt}
\bibliographystyle{unsrtnat}
\bibliography{references}
}

\end{document}